\documentclass[11pt]{article}

\usepackage[final]{acl}

\usepackage{times}
\usepackage{latexsym}

\usepackage[T1]{fontenc}

\usepackage[utf8]{inputenc}

\usepackage{microtype}

\usepackage{inconsolata}

\usepackage{graphicx}

\usepackage{amsmath}
\usepackage{booktabs}
\usepackage{multirow}
\usepackage{multicol}
\usepackage{tcolorbox}
\DeclareSymbolFont{extraup}{U}{zavm}{m}{n}
\DeclareMathSymbol{\varheart}{\mathord}{extraup}{86} 

\tcbset{
  mybox/.style={
    colback=white, colframe=black,
    fonttitle=\bfseries,
    left=1mm, right=1mm, top=1mm, bottom=1mm,
    boxsep=1mm, arc=1mm,
  }
}

\title{Chain-of-thought Reviewing and Correction for \\ Time Series Question Answering}

\author{%
  Chen Su$^{{\spadesuit}}$, \hspace{0.1cm}
    Yuanhe Tian$^{\varheart}$, \hspace{0.1cm}
    Yan Song$^{{\spadesuit}*}$
    \\
    $^{\spadesuit}$University of Science and Technology of China \\
    $^{\varheart}$Zhongguancun Institute of Artificial Intelligence
    \\
    $^{\spadesuit}$\texttt{suchen4565@mail.ustc.edu.cn} \hspace{0.1cm}
    $^{\varheart}$\texttt{tianyuanhe@zgci.ac.cn} \hspace{0.1cm}
    $^{\spadesuit}$\texttt{clksong@gmail.com}
}

\begin{document}
\maketitle
\renewcommand{\thefootnote}{\fnsymbol{footnote}}
\footnotetext[1]{Corresponding author.}
\renewcommand{\thefootnote}{\arabic{footnote}}
\begin{abstract}
With the advancement of large language models (LLMs), diverse time series analysis tasks are reformulated as time series question answering (TSQA) through a unified natural language interface.
However, existing LLM-based approaches largely adopt general natural language processing techniques and are prone to reasoning errors when handling complex numerical sequences.
Different from purely textual tasks, time series data are inherently verifiable, enabling consistency checking between reasoning steps and the original input.
Motivated by this property, we propose T3LLM, which performs multi-step reasoning with an explicit correction mechanism for time series question answering.
The T3LLM framework consists of three LLMs, namely, a worker, a reviewer, and a student, that are responsible for generation, review, and reasoning learning, respectively.
Within this framework, the worker generates step-wise chains of thought (CoT) under structured prompts, while the reviewer inspects the reasoning, identifies erroneous steps, and provides corrective comments.
The collaboratively generated corrected CoT are used to fine-tune the student model, internalizing multi-step reasoning and self-correction into its parameters.
Experiments on multiple real-world TSQA benchmarks demonstrate that T3LLM achieves state-of-the-art performance over strong LLM-based baselines.\footnote{Code available at \url{https://github.com/synlp/T3LLM}.}
\end{abstract}

\begin{figure*}[th]
    \centering
    \includegraphics[width=1.\linewidth, trim=0 20 0 15]{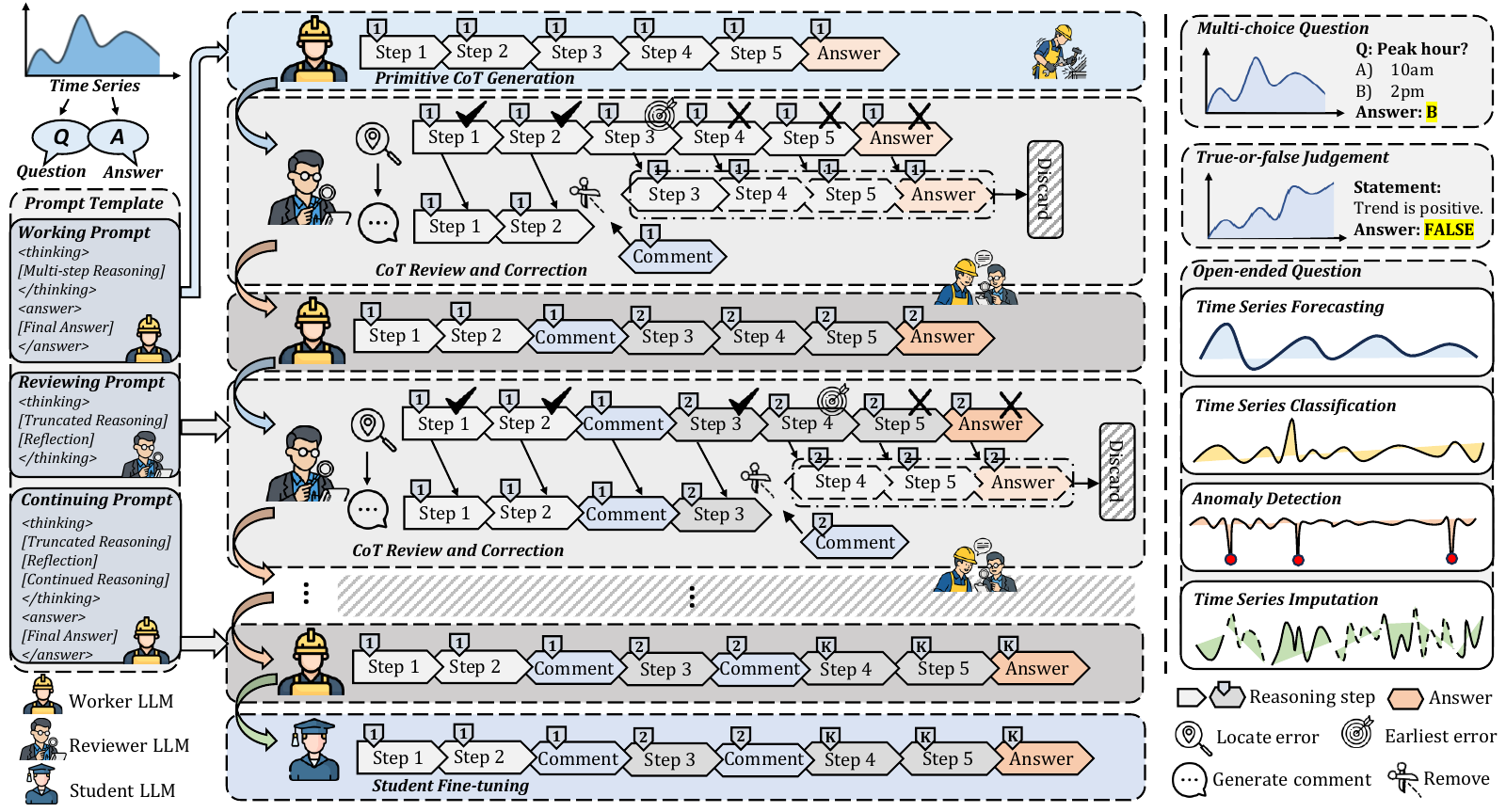}
    \caption{
    The illustration of our approach.
    The left side defines the TSQA task and the working, reviewing, and continuing prompt templates.
    The middle depicts the overall process of T3LLM.
    From top to bottom, they represent the primitive CoT generation, the correction loop, and the fine-tuning based on the CoTs.
    The right side shows the three types of TSQA tasks (i.e., multi-choice questions, true-or-false judgments, and open-ended questions).
    }
    \label{fig:main model}
    \vspace{-0.3cm}
\end{figure*}

\section{Introduction}

Time series analysis (TSA) is a conventional practical task for data and pattern analysis and has undergone the evolution from statistical models \cite{newbold1983arima, de2011forecasting, mondal2014study, liu2016online} to deep learning approaches \cite{lai2018modeling, han2019review, lim2021time, zhou2021informer, liu2022scinet, nie2022time, su2025diffusion, su2025fusing},
which reflects both the growing complexity of temporal data and the continuous innovation in methods to uncover meaningful insights and forecasts in this area \cite{jin2023spatio, liu2024time, su2025multimodal, su2025text}.
As a representative task of TSA,
time series question answering (TSQA) frames problems in time series data as natural language questions and answers, and solves time-related queries within a unified question answering (QA) form.
In TSQA instances, a question usually corresponds to a TSA task such as time series forecasting, imputation, classification, or anomaly detection \cite{wang2025chattime, wang2025itformer, chen2025mtbench},
and the answer is often insufficient with only numerical information, where it is also required to have textual explanations of the reasoning process and key evidence in order to make reliable decisions.

With large language models (LLMs) emerged as universal tools for reasoning across domains \cite{xue2023promptcast, huang2024vtimellm, shu2025audio}, researchers start to explore their applications in TSA since 
text and time series data share a common property of being one-dimensional linear chain structures \cite{zhou2023one, jin2023time, ansari2024chronos}.
So that it
maintains a standardized paradigm that seamlessly aligns with the nature of LLMs and is able to unlock their powerful semantic reasoning capabilities for TSA.
Existing LLM-based TSA approaches \cite{rasul2023lag, jin2023time, luo2025time} are broadly fall into two categories.
The first introduces an independent encoder outside the LLM,
which outputs time series representations and fuse them with text representations for LLM input \cite{jin2023time, hu2025context, zhao2025enhancing}.
Such encoding paradigm easily leads to mismatched semantic spaces across encoders and LLMs.
The other converts the time series data directly into the textual input \cite{xue2023promptcast, ansari2024chronos, luo2025time},
and some studies \cite{kojima2022large, sun2023test}
thus discretizing their values or numeric sequences into textual tokens,
so that allowing LLMs to model TSA tasks within natural language processing (NLP) techniques \cite{ansari2024chronos, chow2024towards, wang2025can, luo2025time},
especially utilizing QA as a direct interface for information retrieval in TSA, even though with simple solutions \cite{xie2024chatts, wang2025chattime, kong2025time}.
However, the widely adopted textualizing solution of long numeric sequences makes the model sensitive to subtle numerical fluctuations when handling complex questions.
To avoid falling into such scenario, a reasonable attempt is to enhance the reasoning capabilities of LLMs, enabling them to break down a task into smaller and manageable sub-tasks \cite{wei2022chain, zhou2022least, sprague2024cot},
therefore, driving some studies to use chain-of-thought (CoT) in TSA to force LLMs to explicitly generate intermediate reasoning steps and prove its effectiveness accordingly \cite{chow2024towards, wang2025can, luo2025time}.
Yet, they still omit the unique characteristic in TSA that the time series information (especially their numerical values) are able to be validated for checking the consistency between the reasoning steps and the original input,
which is different from pure text reasoning.
To this end, strategically tailoring LLMs' utilization to the nature of time series data is essential in applying LLMs to TSA, especially TSQA.
In this paper, we propose an enhanced CoT framework named T3LLM for TSQA, which explicitly utilize a self-corrected reasoning process.
In our framework, three LLMs are assigned with distinct roles, a worker, a reviewer, and a student, respectively.
We adopt a unified CoT prompt that guides the worker model to decompose the time series and question into step-wise reasoning steps, producing primitive chains of thought.
The reviewer model inspects the primitive CoT, identifies the earliest incorrect reasoning step, retains only the verified correct steps, and generates corresponding comments.
The worker model then continues the reasoning based on the retained correct steps and the comments to update the CoT.
This review-and-continuation process is repeated until the reasoning is correct or a predefined iteration limit is reached.
Finally, we collect all output corrected CoTs from the reviewer LLM and utilize them to fine-tune the student LLM so that it is implicitly enhanced with CoT training and thus able to produce logically consistent multi-step reasoning for TSQA.
Extensive experiments on two representative TSQA benchmark datasets demonstrate that T3LLM achieves the state-of-the-art performance
compared with other LLM-based approaches.

\section{The Approach}

Given a time series data $\mathcal{T}$,
and
a set of questions $\mathcal{Q}=\{{Q}_1,{Q}_2,\dots,{Q}_N\}$ with their corresponding answers $\mathcal{A}=\{{A}_1,{A}_2,\dots,{A}_N\}$ derived from $\mathcal{T}$,
TSQA predicts an answer $\widehat{A}_n$ based on a question $Q_n \in \mathcal{Q}$ and the corresponding $\mathcal{T}$.
To perform TSQA, the proposed T3LLM builds a TSQA-specific LLM with multi-step reasoning and review-correction procedure,
with the overall framework illustrated in the Figure \ref{fig:main model}.
Specifically, the T3LLM framework contains three models with different roles, namely, a worker LLM $f_{{\text{w}}}$, a reviewer LLM $f_{{\text{r}}}$, and a student LLM $f_{{\text{s}}}$.
The worker LLM takes an individual question and its corresponding time series as input, producing a step-wise structured primitive CoT and an initial predicted answer.
The reviewer LLM examines the primitive CoT, identifies the earliest incorrect reasoning step, removes this step and all subsequent reasoning, and generates comments for correction.
Guided by the comments, the worker LLM continues generating new reasoning steps, resulting in an updated CoT and a new predicted answer.
The generation and review processes alternate iteratively, forming a progressive correction loop until a predefined stopping criterion is met.
The high-quality corrected CoT collected during this process is used to fine-tune the student LLM.
After fine-tuning, the student LLM is able to produce logically consistent multi-step reasoning without relying on the reviewer and is used as the final model for TSQA.
In the following text, we present the details of the process.

\subsection{Primitive CoT Generation}
In the primitive CoT generation stage, we adopt a working prompt $\mathcal{P}_W$ to regulate the reasoning structure and output format of the worker LLM.
The working prompt guides the worker LLM to generate a structured primitive CoT based on a given question and its corresponding time series.
The prompt introduces a set of predefined special tokens to explicitly separate reasoning content from final answer content.
One category of special tokens (e.g., \texttt{<think>...</think>}) indicates the region in which the model produces step-wise reasoning.
Another category of special tokens (e.g., \texttt{<answer>...</answer>}) indicates the region reserved for generating the final predicted answer.
Under this unified prompt template, the primitive CoT is organized as a sequence of discrete and ordered reasoning steps which is formulated as
\begin{equation}
\setlength{\abovedisplayskip}{5pt}
\setlength{\belowdisplayskip}{5pt}
\mathcal{R}^{(0)} = \big[{R}^{(0)}_{1},{R}^{(0)}_{2},\dots,{R}^{(0)}_{L}\big]
\end{equation}
where $\big[\cdot\big]$ denotes the concatenation of reasoning steps, $L$ denotes the length of the CoT, and ${R}^{(0)}_{l}$ is the $l$-th CoT step.
At this stage, the worker LLM generates the primitive CoT together with an initial predicted answer conditioned on the question, the time series, and the generation prompt.
This step-wise reasoning formulation provides a clear and systematic structural basis for subsequent reasoning review and fine-grained error localization.

\subsection{CoT Review and Correction}

In the CoT review and correction stage, we employ the reviewer LLM to conduct a step-by-step inspection of the primitive CoT generated by the worker LLM, with the goal of identifying the earliest reasoning error.
Specifically, the reviewer LLM takes as input the question $Q_n$, the time series $\mathcal{T}$, the golden standard answer $A_n$, the primitive CoT $\mathcal{R}^{(0)}$, the corresponding predicted answer $\widehat{{A}}_n^{(0)}$, and a reviewing prompt $\mathcal{P}_{R}$, and examines the reasoning steps in a sequential manner.
The reviewer LLM is prompted to locate the earliest reasoning step that conflicts with the time series evidence or the task definition.
Once the erroneous reasoning step is identified, the reviewer LLM removes this step together with all subsequent reasoning steps, retaining only the prefix consisting of verified correct reasoning.
The reviewer LLM then generates a comment that highlights the issues in the current reasoning and provides guidance for correction, without revealing any information about the golden standard answer.
This comment is appended to the truncated CoT, forming an intermediate but corrected reasoning state, which is formulated as
\begin{equation}
\setlength{\abovedisplayskip}{5pt}
\setlength{\belowdisplayskip}{5pt}
\widetilde{\mathcal{R}}^{(1)} = f_{\text{r}}\big(Q_n,\mathcal{T},A_n,\mathcal{R}^{(0)},\widehat{{A}}_n^{(0)},\mathcal{P}_{R}\big)
\end{equation}
where $\widetilde{\mathcal{R}}^{(1)}$ denotes the truncated CoT augmented with the reviewer-generated comment.
It is important to note that this intermediate CoT is still incomplete, and the remaining reasoning steps need to be regenerated by the worker LLM under the guidance of the appended comment.
In the continuation stage, we use a continuing prompt $\mathcal{P}_{C}$ to instruct the worker LLM to generate new reasoning steps and an updated predicted answer conditioned on the truncated CoT.
This review-truncate-comment-continue process is repeated until a predefined maximum number of correction rounds is reached.
For different types of time series question answering tasks, the best corrected CoT is selected based on answer correctness or proximity to the golden standard answer.
The final selected corrected CoT is used as a stable, consistent and high-quality target supervision signal for fine-tuning the student LLM.

\subsection{Student Fine-tuning}

In the student fine-tuning stage, training is performed using individual corrected CoT samples as supervision.
For each training instance, the student model is conditioned on the question, the time series, and the corresponding corrected CoT.
The student model is trained to autoregressively predict the next token in the CoT given the preceding reasoning context.
The CoT loss $\mathcal{L}_{\text{cot}}$ is computed accordingly following the standard cross-entropy loss function.
After generating the CoT, the student model further predicts the token sequence of the final answer conditioned on the completed reasoning, where an answer prediction loss $\mathcal{L}_{\text{ans}}$ following the standard process.
The overall training objective of the student model is to jointly minimize the CoT loss and the answer prediction loss as follows:
\begin{equation}
\setlength{\abovedisplayskip}{5pt}
\setlength{\belowdisplayskip}{5pt}
\mathcal{L} = \mathcal{L}_{\text{cot}} + \mathcal{L}_{\text{ans}}
\end{equation}
By repeating this process over all training instances, the student model internalizes the multi-step reasoning structure and becomes capable of independently generating consistent reasoning for TSQA.

\section{Experiment Settings}

\subsection{Datasets}
We use two datasets in our experiments, namely, CTQA \cite{wang2025chattime} and TMQA \cite{kong2025time}, spanning multiple application domains such as finance, weather, and healthcare, etc.
Both datasets consist of question–answer pairs in which the associated time series information appears in the texts.
Specifically, CTQA contains multiple-choice question (MCQ) and answer pairs,
which are constructed according to trend, volatility, seasonality, and outlier patterns in time series data.
TMQA includes three QA types, namely, MCQ, true-or-false (T/F) judgment and open ended (OPE) QA pairs,
which are constructed from typical TSA tasks, such as time series forecasting, imputation, classification, and anomaly detection.
The time series information used in the QA pairs comes from two sources: public real-world time-series data and synthesized ones from LLMs.
Each question in this dataset additionally provides contextual information such as domain background and descriptions for time series data.
Statistics of these two datasets
according to QA types
are reported in Table \ref{tab:tsqa_stats}.

\begin{table}[t]
\centering
\small
\resizebox{\linewidth}{!}{
\begin{tabular}{l|l|l|rrr}
\toprule
Dataset & \multicolumn{2}{c|}{Type} & Train & Val & Test \\
\midrule
CTQA & MCQ & -- & 26,880 & 6,720 & 14,400 \\
\midrule
\multirow{6}{*}{TMQA}
& MCQ & -- & 6,352 & 1,589 & 3,332 \\
\cmidrule{2-2} \cmidrule{3-3} \cmidrule{4-6}
& T/F & -- & 3,828 & 957 & 2,131 \\
\cmidrule{2-2} \cmidrule{3-3} \cmidrule{4-6}
& \multirow{4}{*}{OPE} & Classification & 20,719 & 5,180 & 11,101 \\
&                     & Anomaly        & 20,720 & 5,180 & 11,100 \\
&                     & Forecasting    & 23,831 & 5,958 & 12,768 \\
&                     & Imputation     & 21,647 & 5,412 & 11,598 \\
\midrule
Total & \multicolumn{2}{c|}{--} & 123,978 & 30,995 & 66,430 \\
\bottomrule
\end{tabular}
}
\vspace{-0.2cm}
\caption{Statistics of the experiment datasets. MCQ, T/F, and OPE stand for multiple-choice, true-or-false, and open-ended QA types, respectively. 
The open-ended QA further includes time series classification, anomaly detection, forecasting, and imputation tasks.}
\label{tab:tsqa_stats}
\vspace{-0.3cm}
\end{table}

\begin{table*}[t]
\centering
\resizebox{\linewidth}{!}{
\begin{tabular}{l|c|cc|ccccccccc}
\toprule
\multirow{4}{*}{\textbf{Approach}}
& \multicolumn{1}{c|}{\textbf{CTQA}}
& \multicolumn{10}{c}{\textbf{TMQA}} \\
\cmidrule(lr){2-2} \cmidrule(lr){3-12}
& \multicolumn{1}{c|}{\multirow{2}{*}{\textbf{MCQ}}}
& \multicolumn{1}{c}{\multirow{2}{*}{\textbf{MCQ}}}
& \multicolumn{1}{c|}{\multirow{2}{*}{\textbf{T/F}}}
& \multicolumn{8}{c}{\textbf{OPE}} \\
\cmidrule(lr){5-12}
& & & &
\multicolumn{2}{c}{\textbf{Classification}}
& \multicolumn{2}{c}{\textbf{Anomaly}}
& \multicolumn{2}{c}{\textbf{Forecasting}}
& \multicolumn{2}{c}{\textbf{Imputation}} \\
\cmidrule(lr){2-2} \cmidrule(lr){3-3} \cmidrule(lr){4-4} \cmidrule(lr){5-6} \cmidrule(lr){7-8} \cmidrule(lr){9-10} \cmidrule(lr){11-12}
& \textbf{Acc} $\uparrow$
& \textbf{Acc} $\uparrow$
& \textbf{Acc} $\uparrow$
& \textbf{Acc} $\uparrow$ & \textbf{mF1} $\uparrow$
& \textbf{Acc} $\uparrow$ & \textbf{F1} $\uparrow$
& \textbf{RMSE} $\downarrow$ & \textbf{MAE} $\downarrow$
& \textbf{RMSE} $\downarrow$ & \textbf{MAE} $\downarrow$ \\
\midrule
TSEnc & 0.399 & 0.483 & 0.518 & 0.320 & 0.274 & 0.286 & 0.246 & 24682.469 & 943.354 & 4668.367 & 94.359 \\
TSTxt  & 0.365 & 0.279 & 0.539 & 0.387 & 0.257     & 0.398   & 0.368    & 15273.367    & 666.316    & 3008.981    & 75.715  \\
TSCoT & 0.390 & 0.574 & 0.635 & \underline{0.495} & 0.293   & 0.513 & 0.393      & 12321.533 & 640.241 & 2900.958 & 72.611  \\
TSCoT+ & 0.393 & \underline{0.589} & 0.629 & 0.483 & \underline{0.317}   & \underline{0.526} & 0.405      & 12359.176 & 658.964 & 2887.367 & 70.851  \\
\midrule
ChatTS        & \underline{0.467} & 0.556 & \underline{0.637} & 0.393 & 0.266 & 0.313 & 0.299 & 26543.138 & 934.681 & 4356.167 & 90.348 \\
ChatTime      & 0.366 & 0.546 &  0.619 & 0.439 & 0.271 & 0.489 & \underline{0.417} & \underline{11926.354} & \underline{628.367} & \underline{2791.376}0 & \underline{64.913} \\
$\text{Time-MQA}^\dagger$       & 0.371 & 0.483 & 0.564 & 0.478 & 0.287 & 0.432 & 0.371 & 17659.672 & 776.128 & 3122.176 & 78.911 \\
\midrule
\midrule
T3LLM & \textbf{0.665} & \textbf{0.635} & \textbf{0.766} & \textbf{0.572} & \textbf{0.336} & \textbf{0.532} & \textbf{0.494} & \textbf{7586.280} & \textbf{545.337} & \textbf{1320.905} & \textbf{50.204} \\
\bottomrule
\end{tabular}
}
\vspace{-0.2cm}
\caption{
Performance comparison of baselines and existing TSQA studies on the benchmark datasets. 
For the MCQ and T/F QAs, the accuracy is reported.
For the classification in OPE, we report accuracy and macro-F1 (mF1).
For the anomaly detection in OPE, we report accuracy and F1.
For the forecasting and imputation in OPE, we report RMSE and MAE.
\textbf{Boldface} and \underline{underlines} denote the best and second-best results, respectively, across all baselines. ``$^\dagger$'' indicates the results of our implementation of the corresponding approach following the original paper.
}
\label{tab:overall_results}
\vspace{-0.3cm}
\end{table*}

\subsection{Baselines and Comparing Approaches}

To evaluate the effectiveness of our proposed T3LLM, we construct four baselines on the same backbone as that used in T3LLM, namely TSEnc, TSTxt, TSCoT, and TSCoT+,
which represent three representative LLM-based approaches to TSQA.
Specifically, TSEnc incorporates an independent time series encoder,
with the output representations concatenated with those from texts, and then fed into the LLM to generate the answer.
This baseline is used for evaluating the performance of TSQA with separate multimodal modeling.
TSTxt treats time series data as normal texts and utilizes the standard procedure of using LLM on them,
which is to test how text-based processing affects TSQA performance.
TSCoT applies CoT to TSTxt with
the same generating prompt as T3LLM, driving the LLM to produce multi-step reasoning for answer generation.
TSCoT+ is a variant of TSCoT that uses a more complex generating prompt with a reflection hint to encourage the LLM to automatically reflect on the previous reasoning steps at an appropriate time.
This baseline is employed to evaluate the ability to utilize standard CoT in TSQA and provides a direct contrast to T3LLM.

To further demonstrate the superiority of our approach, we compare three existing LLM-based TSQA studies under the same experiment setting, namely ChatTS \cite{xie2024chatts}, ChatTime \cite{wang2025chattime} and Time-MQA \cite{kong2025time}.
ChatTS encodes time series into continuous representations and inserts them into text token sequences, thus preserving temporal context information during language modeling.
ChatTime treats discretized time series values as foreign language tokens so that the numerical sequence and text are modeled jointly as if they are in a code-mixture language environment.
Time-MQA directly fine-tunes LLMs with LoRA \cite{hu2022lora} on TSQA data under the standard NLP paradigm, thereby enabling the model to perform TSQA tasks.

\subsection{Implementation Details}
\label{sec:imp}

In data preprocessing, we follow the formatting method used in prior studies \cite{xie2024chatts, kong2025time, wang2025can},
where numerical time series values are kept
and accommodated inside square brackets ``\texttt{[...]}''.\footnote{It is worth noting that, although there are approaches \cite{wang2025chattime} normalize time series values into a unified scale,
we still keep such values in their original scale, which allows the model to directly sense the detailed numerical changes on level, slope, and local fluctuations.}
The working, reviewing, and continuing prompts (i.e., $\mathcal{P}_W$, $\mathcal{P}_R$, and $\mathcal{P}_C$) used in our approach are presented in Appendix \ref{app:prompt_templates}.
Since input representation plays an essential role in data processing \cite{mikolov2013efficient,han2018hyperdoc2vec,devlin2019bert,diao2021tilgan,touvron2023llama,qwen2025qwen25technicalreport}, we utilize state-of-the-art LLMs in the experiments.
Specifically, for the worker and reviewer LLM, we use DeepSeek-R1\footnote{\url{https://huggingface.co/deepseek-ai/DeepSeek-R1}} \cite{guo2025deepseek}
For the student LLM, we employ Qwen2.5-14B-Instruct\footnote{\url{https://huggingface.co/Qwen/Qwen2.5-14B-Instruct}}.
We fine-tune the student LLM using LoRA \cite{hu2022lora} with LoRA rank {64}, LoRA alpha {128} and LoRA dropout {0.05}.
We use the AdamW optimizer \cite{loshchilov2017decoupled} with learning rate {2e-5}, and {3} training epochs.
We set the maximum CoT length as {2048} tokens
and the maximum correction round ($\mathcal{MCR}$) as {3}
to avoid excessively long reasoning,
In evaluation, we use specific metrics for different QA types.
Particularly, for MCQ, T/F, and classification and anomaly detection tasks in OPE QA, we parse the generated answers into discrete labels and compute accuracy and macro-averaged F1 (mF1) accordingly. For imputation and forecasting tasks in open-ended QA, we compute mean absolute error (MAE) and root mean squared error (RMSE) on the generated time series values extracted directly from the final answer text.
\section{Results and Analysis}
\subsection{Overall Results}

Table \ref{tab:overall_results} reports the results of different approaches,
with several observations.
First, TSEnc performs well on both MCQ and T/F questions, as well as
the classification and anomaly detection tasks in OPE QA types,
as compared with inferior results in time series forecasting and imputation tasks.
This observation indicates that an independent time series encoder produces rather reliable representations and thus
makes the TSEnc effective on comprehension QA, such as MCQ and T/F ones,
similar to the practice of utilizing better representations in natural language understanding cases.
However, 
since different encoders may vary in their semantic spaces,
TSEnc thus performs poorly when it is required to generate the time series values corresponding to the question.
Second, TSTxt shows a more balanced performance compared to TSEnc,
where 
its performance in time series forecasting and imputation in OPE QA is better than TSEnc, and others are worse.
It suggests that jointly modeling time series values and text in a unified text processing method is effective 
when answering questions that require reasoning over both the time series and text.
Third, TSCoT and TSCoT+ improve TSTxt on comprehension QA as well as time series forecasting and imputation tasks in OPE QA,
which confirms that CoT strengthens time series understanding of LLMs.
Meanwhile, we observe that the performance of TSCoT+ is similar to that of TSCoT, indicating that employing prompt only is insufficient to enable effective reflections.
Compared to the three baselines, T3LLM achieves the best performance across all QA types,
indicating that incorporating enhanced CoT on the basis of time series textualization and CoT mechanism strengthens reasoning, which proves the validity of validating the reasoning process with time series evidence.

Table \ref{tab:overall_results} also reports the results of existing representative LLM-based TSQA approaches.
We observe that ChatTS performs well in MCQ, T/F QA types, and the classification, anomaly detection tasks in OPE questions.
This observation confirms again that using a separate encoder for time series inputs is an effective solution for comprehension QA tasks,
which corresponds to that presented by TSEnc.
In contrast, ChatTime and Time-MQA offer better performance in forecasting and imputation tasks,
which also corresponds to TSTxt, TSCoT, and TSCoT+ in the way that unifying time series and text modeling has the advantage of 
{being able to perceive the most original input values.
Furthermore, ChatTime performs better
than Time-MQA,
which is because ChatTime learns time series values as foreign language tokens, so that simplifies the process of aligning time series information with texts, other than directly mixing them in LLM fine-tuning.
Similar to the comparison with baseline models, T3LLM also achieves the best performance in all QA types than existing studies,
confirming its effectiveness with a carefully tailored CoT training process.

\begin{figure}[t]
    \centering
    \includegraphics[width=\linewidth, trim=0 15 0 0]{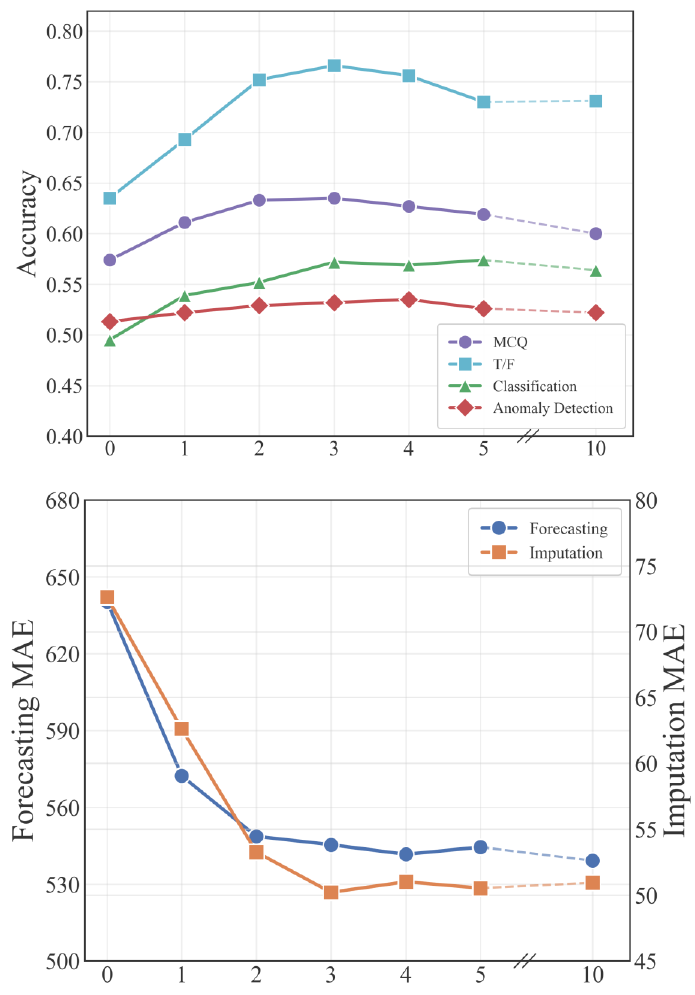}
    \caption{
    The performance of our approach with different numbers of maximum correction rounds $\mathcal{MCR}=\{1,2,3,4,5,10\}$ on the TMQA dataset. 
    }
    \label{fig:effect_of_correction_round}
    \vspace{-0.3cm}
\end{figure}

\begin{table*}[t]
\centering
\resizebox{\linewidth}{!}{
\begin{tabular}{l|cc|cccccccc}
\toprule
\multirow{3}{*}{\textbf{Reviewer LLM}}
& \multirow{2}{*}{\textbf{MCQ}} & \multirow{2}{*}{\textbf{T/F}}
& \multicolumn{8}{c}{\textbf{OPE}} \\
 \cmidrule(lr){4-11}
& & & \multicolumn{2}{c}{\textbf{Classification}}
& \multicolumn{2}{c}{\textbf{Anomaly}}
& \multicolumn{2}{c}{\textbf{Forecasting}}
& \multicolumn{2}{c}{\textbf{Imputation}} \\
\cmidrule(lr){2-2} \cmidrule(lr){3-3}
\cmidrule(lr){4-5} \cmidrule(lr){6-7} \cmidrule(lr){8-9} \cmidrule(lr){10-11}
& \textbf{Acc}$\uparrow$ & \textbf{Acc}$\uparrow$
& \textbf{Acc}$\uparrow$ & \textbf{mF1}$\uparrow$
& \textbf{Acc}$\uparrow$ & \textbf{F1}$\uparrow$
& \textbf{RMSE}$\downarrow$ & \textbf{MAE}$\downarrow$
& \textbf{RMSE}$\downarrow$ & \textbf{MAE}$\downarrow$ \\
\midrule
DeepSeek-R1            & \textbf{0.635} & \textbf{0.766} & \textbf{0.572} & \textbf{0.336} & \textbf{0.532} & \textbf{0.494} & \textbf{7586.280}  & \textbf{545.337} & \textbf{1320.905} & \textbf{50.204} \\
DeepSeek-V3            & 0.603 & 0.680 & 0.539 & 0.325 & 0.530 & 0.445 & 8934.836  & 588.148 & 2141.449 & 58.174 \\
Qwen2.5-14B-Distill    & 0.594 & 0.661 & 0.518 & 0.312 & 0.528 & 0.439 & 9926.251  & 617.422 & 2473.362 & 62.492 \\
Qwen2.5-14B-Instruct   & 0.589 & 0.648 & 0.499 & 0.309 & 0.524 & 0.434 & 11874.432 & 622.892 & 2843.963 & 66.748 \\
\bottomrule
\end{tabular}

}
\vspace{-0.2cm}
\caption{
Performance comparison of different reviewer LLMs on the TMQA dataset.
}
\label{tab:effect_of_different_reviewer_llm}
\vspace{-0.3cm}
\end{table*}

\begin{table*}[t]
\centering
\resizebox{\linewidth}{!}{
\begin{tabular}{l|cc|cccccccc}
\toprule
\multirow{3}{*}{\textbf{Worker LLM}}
& \multirow{2}{*}{\textbf{MCQ}} & \multirow{2}{*}{\textbf{T/F}}
& \multicolumn{8}{c}{\textbf{OPE}} \\
 \cmidrule(lr){4-11}
& & & \multicolumn{2}{c}{\textbf{Classification}}
& \multicolumn{2}{c}{\textbf{Anomaly}}
& \multicolumn{2}{c}{\textbf{Forecasting}}
& \multicolumn{2}{c}{\textbf{Imputation}} \\
\cmidrule(lr){2-2} \cmidrule(lr){3-3}
\cmidrule(lr){4-5} \cmidrule(lr){6-7} \cmidrule(lr){8-9} \cmidrule(lr){10-11}
& \textbf{Acc}$\uparrow$ & \textbf{Acc}$\uparrow$
& \textbf{Acc}$\uparrow$ & \textbf{mF1}$\uparrow$
& \textbf{Acc}$\uparrow$ & \textbf{F1}$\uparrow$
& \textbf{RMSE}$\downarrow$ & \textbf{MAE}$\downarrow$
& \textbf{RMSE}$\downarrow$ & \textbf{MAE}$\downarrow$ \\
\midrule
DeepSeek-R1            & \textbf{0.635} & \textbf{0.766} & \textbf{0.572} & \textbf{0.336} & \textbf{0.532} & \textbf{0.494} & \textbf{7586.280}  & \textbf{545.337} & \textbf{1320.905} & \textbf{50.204} \\
DeepSeek-V3            & 0.624 & 0.742 & 0.566 & 0.331 & 0.529 & 0.490 & 7835.424  & 566.142 & 1533.242 & 51.821 \\
Qwen2.5-14B-Distill    & 0.591 & 0.639 & 0.525 & 0.324 & 0.525 & 0.474 & 8439.348  & 574.334 & 2214.422 & 64.334 \\
Qwen2.5-14B-Instruct   & 0.573 & 0.618 & 0.496 & 0.248 & 0.520 & 0.432 & 12421.224 & 624.422 & 2794.996 & 70.382 \\
\bottomrule
\end{tabular}
}
\vspace{-0.2cm}
\caption{
Performance comparison of different worker LLMs on the TMQA dataset.
}
\label{tab:effect_of_different_worker_llm}
\vspace{-0.3cm}
\end{table*}

\subsection{Effect of Different $\mathcal{MCR}$}

To investigate the effect of different maximum correction rounds $\mathcal{MCR}$, we conduct an analysis with $\mathcal{MCR}$ settings of $1, 2, 3, 4, 5, 10$.
with
the results shown in Figure \ref{fig:effect_of_correction_round}.
As a comparison, we also plot TSCoT's results as the standard CoT baseline in the same Figure.
We observe that, when $\mathcal{MCR}=1$, T3LLM already significantly outperforms the no-review-correction TSCoT baseline on all TSQA tasks,
and its performance keeps increasing when raising $\mathcal{MCR}$ from $1$ to $3$.
the performance on classification and anomaly detection in OE task} gradually converges with higher $\mathcal{MCR}$.
The reason is that
additional correction rounds provide more help to high-complexity tasks, such as time series forecasting and imputation, and less to the others, so that
smaller gains are expected for the classification and anomaly detection tasks in OPE QA, respectively.
A detailed explanation of this phenomenon is that forecasting and imputation questions often require a larger reasoning space
and more rounds of review and correction help to adjust potential reasoning bias and redirect it to reliable paths.
Still,
when further increasing $\mathcal{MCR}$ to $5$, the performance stays stable,
which suggests that the benefit of larger $\mathcal{MCR}$ saturates all possible enhancements at its earlier stage, so that more rounds do not provide new information to help TSQA.
We also test the case with $10$ maximum correction rounds to explore performance under extremely round settings.
We found that the performance of $10$ setting is even slightly lower than that of $5$ correction rounds.
This indicates that long correction rounds create too many comments, making the LLM give more conservative answers and reducing performance.
As a result, this analysis offers a reasonable trade-off in setting $\mathcal{MCR}$ between performance and computation cost.

\subsection{Effect of Different Reviewer LLMs}  
We evaluate different LLMs as reviewer LLMs in our approach and conduct experiments on the TMQA dataset.
We try DeepSeek-V3\footnote{\url{https://huggingface.co/deepseek-ai/DeepSeek-V3}} \cite{liu2024deepseek}, DeepSeek-R1 \cite{guo2025deepseek}, Qwen2.5-14B-instruct \cite{qwen2025qwen25technicalreport} and Qwen2.5-14B-Distill\footnote{{\url{https://huggingface.co/deepseek-ai/DeepSeek-R1-Distill-Qwen-14B}}},
and keep the DeepSeek-R1 as the worker LLM unchanged in all cases.
{Among these LLMs, Deepseek-R1 has the strongest reasoning ability, followed by Qwen2.5-14B-Distill, while DeepSeek-V3 and Qwen2.5-14B-instruct have relatively weaker reasoning abilities \cite{guo2025deepseek}.}
Results are reported in Table \ref{tab:effect_of_different_reviewer_llm}.
When switching the reviewer LLM to DeepSeek-V3, the obtained student's performance decreases compared to {DeepSeek-R1}, especially in time series forecasting and imputation tasks.
It clearly demonstrates that the reasoning ability of LLMs affects the quality of their corrected CoT, so that delivers to the student's TSQA performance.
In using Qwen2.5-14B-Distill, we observe that T3LLM's performance in MCQ and T/F questions is similar to that with the DeepSeek-V3.
This observation indicates that LLMs with fewer parameters are enough for comprehension QA tasks once they have sufficient reasoning abilities.
In contrast, Qwen2.5-14B-instruct shows further decrease in forecasting and imputation performance, confirming that reasoning ability is inevitable in complex QA scenarios, especially when generation is utilized in OPE questions.

\begin{figure*}[t]
    \centering
    \includegraphics[width=\linewidth, trim=0 15 0 0]{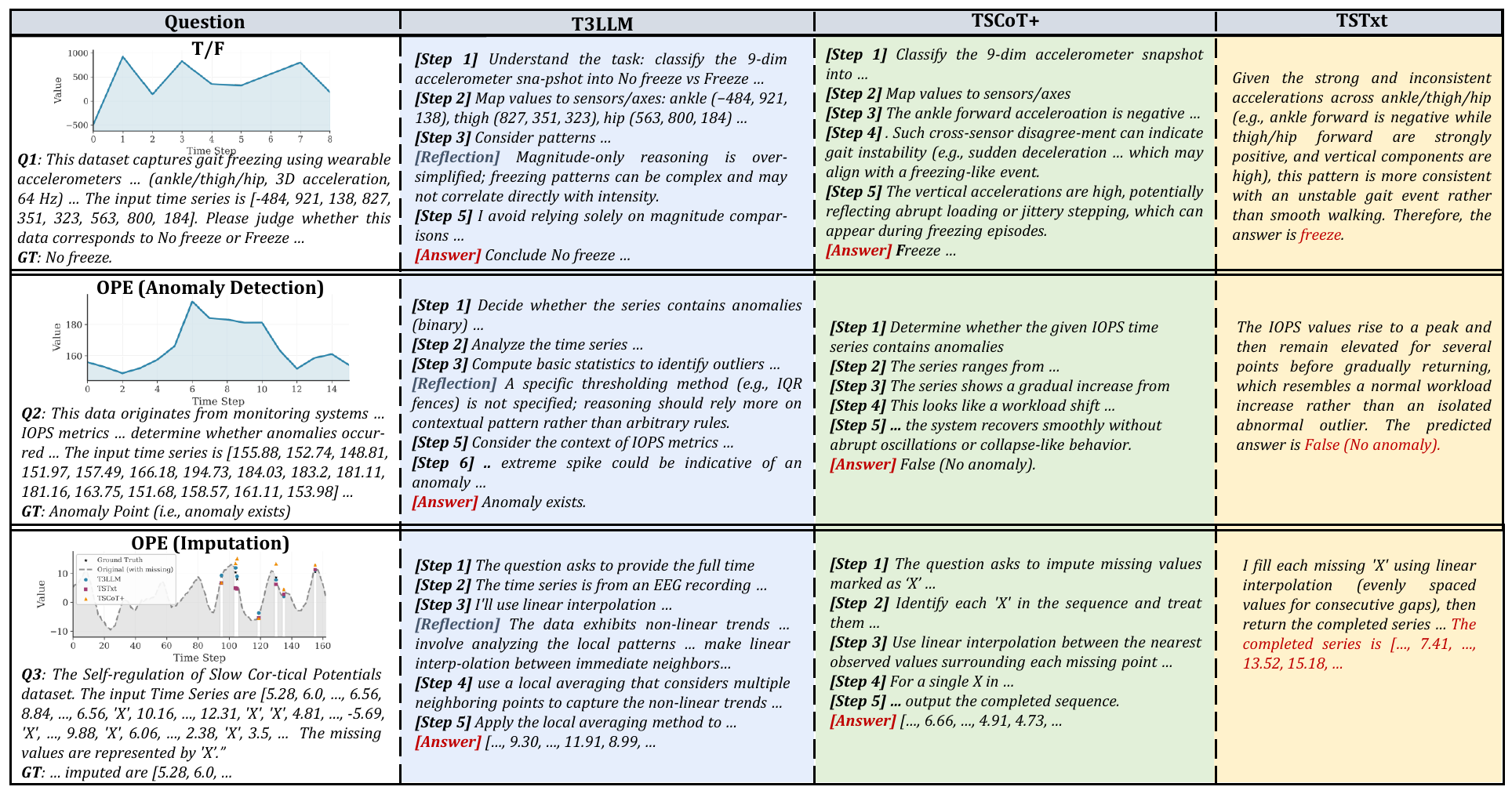}
    \caption{
    Case study with four examples on MCQ and T/F questions, and time series forecasting and classification tasks in OPE questions from the TMQA dataset. 
    The left side presents the question and the corresponding time series.
    The right side show Time-MQA, TSCoT and T3LLM's reasoning process and their corresponding answer.
    }
    \label{fig:case_study}
    \vspace{-0.3cm}
\end{figure*}

\subsection{Effect of Different Worker LLMs}

We further evaluate different LLMs as the worker LLMs in our approach and conduct the experiment on the TMQA dataset.
Similar to the analysis of review LLMs,
we also select DeepSeek-V3, DeepSeek-R1, Qwen2.5-14B-instruct, and Qwen2.5-14B-distill as worker LLMs, while keeping the review LLM fixed with DeepSeek-R1.
The results are reported in Table \ref{tab:effect_of_different_worker_llm}.
We observe that utilizing DeepSeek-V3 as the worker model leads to a substantial performance degradation of T3LLM across all QA tasks,
which indicates that strong reasoning capability is also crucial for the worker LLM in our approach, because the initial CoT is crucial in starting our review and correction process.
When using Qwen2.5-14B-Distill and Qwen2.5-14B-instruct as the worker LLM, the performance 
on all types of question further decreases,
and we find that such smaller LLMs often fail to produce correct CoT solutions in the first place, so the later correction process is unable to stop the student LLM from learning erroneous CoTs.

\subsection{Case Study}

In order to have a more detailed understanding of how T3LLM performs on TSQA, Figure \ref{fig:case_study} visualizes its review and correction process with a few example instances and their generated answers
on the T/F question (Q1) and time series anomaly detection (Q2) and imputation (Q3) tasks in OPE questions.
As a comparison, the figure also shows the responses of TSCoT+ and the baseline TSTxt on the same examples.
Across all cases, T3LLM clearly shows its power on correcting 
critical reasoning steps to resolve inconsistencies, thus yielding more accurate answers.
In contrast, TSTxt often produces limited explanatory texts, so that shows obvious deviations
from ground truth on questions that require fine-grained numerical reasoning.
While TSCoT+ is also able to produce multi-step reasoning, it yet lacks effective self-correction, thus often causes mistakes
in early reasoning steps that lead to subsequent error propagation later.

\section{Related Work}

TSA plays an important role in data and pattern analysis \cite{newbold1983arima, williams1998urban, hyndman2008forecasting, de2011forecasting, liu2016online, han2019review, wen2022transformers, jin2023spatio, liu2024time, su2025multimodal}, with its methodologies spanning
from early statistical approaches such as autoregressive integrated moving average (ARIMA) \cite{box1968some} or exponential smoothing \cite{holt2004forecasting} that 
model the variation patterns
\cite{zhang2003time, hyndman2008forecasting, de2011forecasting, mondal2014study}
to later neural solutions using
Convolutional or recurrent networks that fit nonlinear temporal relationships in time series data \cite{oreshkin2019n, han2019review, lim2021time}.
Recently, the Transformer-based TSA approaches extend the task to longer time series input \cite{zhou2021informer, wen2022transformers, nie2022time},
and transfer to LLM-based approaches \cite{xie2024chatts, alnegheimish2024large, chow2024towards, wang2025chattime, kong2025time, wang2025toward} for this task with the rise of LLMs \cite{wei2022chain,touvron2023llama,lu2023ziya,gan2023idesigner,li2024challenging,tian-etal-2024-chimed,tian2025multimodal,guo2025deepseek,qwen2025qwen25technicalreport}.
Along this line of research, some studies try to maximize the characteristics of time series information by
replacing the input and output layer of LLMs with specific TSA modules \cite{rasul2023lag, zhang2023insight, zhou2023one, das2024decoder, chow2024towards, xie2024chatts, wang2025itformer},
while others seek external help from multimodal information such as text and images in order to provide richer context for TSA tasks \cite{jin2023time, zhong2025time, hu2025context, cheng2025instructime}.
Compared to such studies utilizing additional modules, there are also direct solutions treating time series data as text and using LLMs with their standard language modeling paradigm \cite{ansari2024chronos, jin2024position, wang2025chattime, kong2025time},
or even employing advanced NLP techniques such as 
retrieval augmentation \cite{yang2025timerag, xiao2025retrieval, han2025retrieval}.
and
CoT \cite{wang2025can, luo2025time}, etc.
As a natural task interface, TSQA is proposed to tackle TSA with natural language interactions.
\citet{xie2024chatts} start to construct TSQA data for time series trend prediction and causal inference,
aligning time series data with textual descriptions in order to maintain diverse QA capabilities.
\citet{wang2025chattime} introduces a vocabulary for time-series numerical values, and inserts them in LLM training, mainly focusing on MCQ questions.
\citet{kong2025time} use self-constructed TSQA data to finetune an LLM through varying prompts to handle different tasks such as forecasting, anomaly detection, etc.
They are limited to simple solutions to QA for TSA,
exhibiting poor adaptability to complex time series questions.
Compared to them, T3LLM offers a more reliable framework in utilizing LLMs to mitigate the vulnerability of their operations more effectively with decomposition of reasoning steps and correction of them.

\section{Conclusion}

In this paper, we propose T3LLM,
reviewing and correcting multi-step reasoning to improve LLMs' ability for TSQA.
T3LLM first generates primitive CoTs via a worker LLM, then has them reviewed by a reviewer LLM and continued by the worker, and finally uses these self-corrected CoTs to fine-tune a compact student LLM so that the reasoning process is internalized into the student’s parameters.
Experiments on two TSQA benchmark datasets demonstrate
that T3LLM consistently outperforms existing LLM-based approaches across multiple tasks in TSQA, confirming the validity and effectiveness of reviewing and correction for CoT.



\bibliography{custom}

\appendix

\section{Prompt Templates}
\label{app:prompt_templates}
We provide three complementary prompt templates used in the T3LLM framework: the working template (Figure \ref{fig:working_prompt_template}), the reviewing template (Figure \ref{fig:reviewing_prompt_template}), and the continuing template (Figure \ref{fig:continuing_prompt_template}).

\begin{figure*}[t]
\centering
\begin{tcolorbox}[title=Working Prompt Template, mybox, width=\textwidth]
\footnotesize\ttfamily
System prompt: \\
You are an expert in time series analysis and time series question answering (TSQA). \\
\\
For every query: \\
- Carefully read the question, including the dataset description, task description, label or answer definitions, and the given time series. \\
- Perform explicit step-by-step reasoning ONLY inside the \textless think\textgreater...\textless /think\textgreater\ block. \\
- Ground your reasoning in the numeric time series and the task definitions, not in external knowledge or vague intuition. \\
- After finishing the reasoning, output the final answer ONLY inside the \textless answer\textgreater...\textless /answer\textgreater\ block. \\
- The final answer must strictly follow the format requested in the question (e.g., a single label, one of the given options, True/False, or a list of numbers), and should be as concise as possible. \\
- Do NOT include the final answer text inside \textless think\textgreater. Do NOT repeat the reasoning inside \textless answer\textgreater. \\
\\
User prompt: \\
You are given a time series question answering (TSQA) instance. \\
\\
$[\text{Question}]$ \\
\{add question here\} \\
\\
Your task: \\
1. Understand the task type and the required answer format from the question. \\
2. Carefully analyze the given time series (level, trend, variation, local anomalies, missing values, etc.) together with the context in the question. \\
3. Use step-by-step reasoning, grounded in the time series and task definitions, to decide the correct answer. \\
4. At the end, give your final answer in exactly the format requested by the question (for example: a single label, one of the given options, True/False, or a list of numbers). \\
\\
Output format: \\
\textless think\textgreater \\
$[\text{Step 1}]$ ... \\
$[\text{Step 2}]$ ... \\
$[\text{Step 3}]$ ... \\
$[\text{Step 4}]$ ... \\
(Use as many steps as needed for your reasoning. Do NOT output the final answer here.) \\
\textless /think\textgreater \\
\textless answer\textgreater \\
$[\text{Final answer only, strictly following the required format.}]$ \\
\textless /answer\textgreater
\end{tcolorbox}
\vspace{-0.4cm}
\caption{The working prompt template used in T3LLM.}
\label{fig:working_prompt_template}
\end{figure*}

\begin{figure*}[t]
\centering
\begin{tcolorbox}[title=Reviewing Prompt Template, mybox, width=\textwidth]
\footnotesize\ttfamily
System prompt: \\
You are a rigorous reviewer of time-series reasoning traces. \\
\\
For each case, you will receive: \\
- A question (including the time series and task description), \\
- The ground-truth answer, \\
- A model's previous output containing a reasoning chain in \textless think\textgreater...\textless /think\textgreater\ and a final answer in \textless answer\textgreater...\textless /answer\textgreater. \\
\\
Your job is to: \\
- Decide whether the reasoning chain contains any incorrect, unsupported, or misleading step, given the question, the time series, and the ground-truth answer. \\
- If there is an error, you must locate the FIRST problematic step, insert a self-reflection after that step, and truncate all later steps. You must NOT produce any final answer. \\
- If there is no error and the final answer matches the ground-truth answer, you must NOT modify the reasoning at all and you must NOT reveal or hint at the ground-truth answer. In this case, just output a special marker indicating that no change is needed. \\
\\
Important constraints: \\
- Never reveal or explicitly mention the ground-truth answer in your output. \\
- Never hint which label, option, or numeric values are correct. \\
- Do not output any \textless answer\textgreater\ block. \\
\\
User prompt: \\
$[\text{Question}]$ \\
\{original question\} \\
\\
$[\text{Ground-truth Answer}]$ \\
\{gold answer\} \\
\\
$[\text{Original Model Output}]$ \\
\textless think\textgreater \\
\{working llm's reasoning chain\} \\
\textless /think\textgreater \\
\textless answer\textgreater \\
\{working llm's final answer\} \\
\textless /answer\textgreater \\
\\
Your reviewing task: \\
\\
1. Carefully read the question, the time series, and the ground-truth answer. \\
2. Check whether the final answer in \textless answer\textgreater\ matches the ground-truth answer exactly. \\
3. Read the reasoning steps inside \textless think\textgreater\ in order, from the beginning to the end. \\
4. Identify the FIRST step that is logically wrong, inconsistent with the given time series, inconsistent with the task definitions, or that leads toward an incorrect or unjustified conclusion. \\
\hspace*{2em}- A step is "wrong" if it mis-describes the numeric pattern, misapplies a definition, or supports a conclusion that contradicts the correct answer. \\
5. If you find such an incorrect or problematic step, construct a revised thinking trace as follows: \\
\hspace*{2em}- Copy all correct steps BEFORE the first wrong step without changing their content (you may keep existing step markers like $[\text{Step 1}]$, $[\text{Step 2}]$, etc.). \\
\hspace*{2em}- Copy the first wrong step as well. \\
\hspace*{2em}- Immediately AFTER this wrong step, insert a new line starting with $[\text{Reflection}]$ that: \\
\hspace*{4em}* explicitly explains why the previous step is incorrect, imprecise, or misleading, \\
\hspace*{4em}* briefly indicates in abstract terms what kind of reasoning direction would be more appropriate, WITHOUT revealing or hinting at the ground-truth answer (no explicit labels, options, or exact numeric targets). \\
\hspace*{2em}- DELETE all remaining steps after this reflection and DELETE the final answer. Do NOT continue the reasoning beyond the reflection. \\
6. If you CANNOT find any real error in the reasoning and the final answer already matches the ground-truth answer: \\
\hspace*{2em}- Do NOT modify, repeat, or summarize the original reasoning. \\
\hspace*{2em}- Do NOT output any explanation, reflection, or hint about why it is correct. \\
\hspace*{2em}- Simply output a special marker indicating that no change is needed. \\
Notes: \\
- Never output any \textless answer\textgreater\ block. \\
- Never reveal or hint at the ground-truth answer (no explicit correct label, option, or full numeric sequence).
\end{tcolorbox}
\vspace{-0.4cm}
\caption{The reviewing prompt template used in T3LLM.}
\label{fig:reviewing_prompt_template}
\end{figure*}

\begin{figure*}[t]
\centering
\begin{tcolorbox}[title=Continuing Prompt Template, mybox, width=\textwidth]
\footnotesize\ttfamily
System prompt: \\
You are an expert in time series analysis and time series question answering (TSQA). \\
\\
For every query: \\
- Carefully read the question, including the dataset description, task description, label or answer definitions, and the given time series. \\
- You will receive a partially corrected reasoning trace inside \textless think\textgreater...\textless /think\textgreater. This trace may already contain several steps and one or more $[\text{Reflection}]$ lines that point out previous errors or adjustments. \\
- You must treat the existing content inside \textless think\textgreater\ as fixed. Do NOT modify, delete, or reorder any existing steps or reflections. \\
- Your job is to CONTINUE the reasoning from the end of the given \textless think\textgreater\ block, adding new steps that follow the corrections suggested by the latest $[\text{Reflection}]$. \\
- Explicit step-by-step reasoning must appear ONLY inside the \textless think\textgreater...\textless /think\textgreater\ block. \\
- After finishing the reasoning, output the final answer ONLY inside the \textless answer\textgreater...\textless /answer\textgreater\ block. \\
- The final answer must strictly follow the format requested in the question (e.g., a single label, one of the given options, True/False, or a list of numbers), and should be as concise as possible. \\
- Do NOT include the final answer text inside \textless think\textgreater. Do NOT repeat the reasoning inside \textless answer\textgreater. \\
\\
User prompt: \\
You are given a time series question answering (TSQA) instance together with a partially corrected reasoning trace. \\
\\
$\text{[Question]}$ \\
\{original question\} \\
\\
$\text{[Partially corrected reasoning trace]}$ \\
\textless think\textgreater \\
\{revised reasoning chain\} \\
\textless /think\textgreater \\
\\
Your task: \\
1. Carefully read the question and understand the task type and the required answer format. \\
2. Read the existing reasoning inside \textless think\textgreater. Earlier steps and $[\text{Reflection}]$ lines indicate which directions were previously problematic and how the reasoning should be adjusted in general. \\
3. WITHOUT changing any existing content inside \textless think\textgreater, continue the reasoning by adding new steps AFTER the last existing line inside \textless think\textgreater. \\
\hspace*{2em}- Your new steps should respect the corrections suggested by the latest $[\text{Reflection}]$. \\
\hspace*{2em}- Avoid repeating the same errors that were pointed out by any $[\text{Reflection}]$. \\
4. Based on the completed reasoning, decide the correct final answer in the exact format required by the question. \\
\\
Output format: \\
\textless think\textgreater \\
\{First, copy the entire given reasoning trace EXACTLY as it is, including all previous steps and $[\text{Reflection}]$ lines.\} \\
$[\text{Next Step}]$ ... \\
$[\text{Next Step}]$ ... \\
(... add as many new steps as needed; do NOT include the final answer text here.) \\
\textless /think\textgreater \\
\textless answer\textgreater \\
$\text{[Final answer only, strictly following the required format.]}$ \\
\textless /answer\textgreater
\end{tcolorbox}
\vspace{-0.4cm}
\caption{The continuing prompt template used in T3LLM.}
\label{fig:continuing_prompt_template}
\end{figure*}


\end{document}